# Comparative Analysis of Time Series Forecasting Approaches for Household Electricity Consumption Prediction

**MUHAMMAD BILAL[1], HYEOK KIM[2], MUHAMMAD FAYAZ[3], PRAVIN PAWAR[1]**

[1]Department of Computer Science
State University of New York, Korea
Incheon Global Campus, Incheon, South Korea.
{Muhammad.Bilal, Pravin.Pawar}@stonybrook.edu
[2]Ninewatt Inc.
Incheon Global Campus, Incheon, South Korea.
hyeok0724.kim@gmail.com
[3]Department of Computer Science
University of Central Asia, Naryn Campus
Naryn, Kyrgyz Republic.
Muhammad.Fayaz@ucentralasia.org

**Abstract:** As a result of increasing population and globalization, the demand for energy has greatly risen. Therefore, accurate energy consumption forecasting has become an essential prerequisite for government planning, reducing power wastage and stable operation of the energy management system. In this work we present a comparative analysis of major machine learning models for time series forecasting of household energy consumption. Specifically, we use Weka – a data mining tool to first apply models on hourly and daily household energy consumption datasets available from Kaggle data science community. The models applied are: Multi-Layer Perceptron, K-Nearest Neighbor regression, Support Vector Regression, Linear Regression, and Gaussian Processes. Secondly, we also implemented time series forecasting models - ARIMA and VAR- in python to forecast household energy consumption of selected South Korean households with and without weather data. Our results show that the best methods for the forecasting of energy consumption prediction are Support Vector Regression followed by Multilayer Perceptron and Gaussian Process Regression.

**Keywords:** Time Series Forecasting, ARIMA, VAR, WEKA, Multi-Layer Perceptron, K-Nearest Neighbor regression, Support Vector Regression, Linear Regression, and Gaussian Processes.

## 1. INTRODUCTION

Time Series forecasting is a technique for the prediction of future values based on previously observed values. The technique is used across many fields of study, from weather forecasting to stock market analysis. It uses a model that learns from past data by analyzing the past characteristics of the data and predicts future values under the assumption that the future trends will follow a similar pattern to past trends. Major applications of time series forecasting include economic forecasting, sales forecasting, budgetary analysis, stock market analysis, yield projections, process and quality control, inventory studies, workload projections, utility studies, census analysis and many more [1].

The most commonly methods used for time series forecasting are Multi-Layer Perceptron (MLP), Bayesian Neural Network (BNN), Radial Basis Functions (RBF), Generalized Regression Neural Networks (GRNN) also called kernel regression, K-Nearest Neighbor regression (KNN), CART regression trees (CART), Support Vector Regression (SVR), Gaussian Processes (GP) and recently invented Recurrent Neural Network (RNN) and Long Short-Term Memory (LSTM) models [2].

In this paper, we discuss the results of applying some of these models to forecast household energy consumption as such forecasting is pivotal to understanding energy demand and supply, discovering optimal individual energy usage plans, performing behavioral analysis, performing correlation analysis, studying carbon footprint and much more. Our major motivation for forecasting energy consumption is also to perform behavioral analysis on senior citizens in South Korea and to find an optimal energy usage plan for each house. The most suitable algorithms discovered from this study will be implemented in a cloud computing platform that collects historical electricity usage data from IoT enabled electricity measurement devices, indoor and outdoor temperature as well as real-time weather data and uses this information for electricity usage prediction.

The remainder of this paper is organized as follows. In Section 2 we present the methodologies used by other researchers to perform time series forecasting and brief introduction to the algorithms that we have used in our experiments. In Section 3, we discuss the methodology for our experiments. In Section 4, we present the results of experiments followed by the conclusion.

# 2. RELATED WORK AND ALGORITHMS

## 2.1 Related Work

In recent years, analysis of time series forecasting models has been a very popular research area. The research reported by Seethalakshmi et. al. in [3] uses Long Short Term Memory (LSTM) to predict the energy demand in a smart grid. Similarly, Zang et al. [4] used Autoregressive Moving Average (ARMA) and Autoregressive Integrated Moving Average (ARIMA) to forecast energy consumption. Krishnan et al. [5] looked at predicting energy demand in a smart grid using a combination of (linear) seasonal auto-regressive integrated moving average (SARIMA) and (nonlinear) long-short term memory (LSTM) models. Lee et al. [6] forecasted the energy consumption for a building using several deep learning methods, whereas the LSTM/GRU model showed the best overall performance. Fayaz et. al. [7] proposed a methodology for energy consumption prediction in residential buildings using deep extreme machine learning approach. The methodology proposed in [7] consists of four different layers, namely data acquisition, preprocessing, prediction, and performance evaluation.

Our objective is to predict energy consumption for individual households with senior citizens in South Korea and then perform behavioral analysis using obtained patterns. Towards fulfilling this objective, in this study, we first do a comparative study of different machine learning models available in Weka software [8] on hourly and daily household energy consumption data. Afterwards, we apply two statistical models, ARIMA and VAR, on Korean household data collected by our IoT based cloud computing platform.

## 2.2 Introduction to Time Series Forecasting Algorithms

In this section, we provide a brief introduction to 5 classical models for time series forecasting in Weka and ARIMA and VAR models (which we implemented in Python). The reason for selecting these models is that they are some of the most commonly used models.

### A. Linear Regression

Linear regression is a linear approach to modeling the relationship between the input variables **x** and the single output variable **y**, which can be calculated from a linear combination of the input variables **x**. When there is a single input variable **x**, the method is referred to as simple linear regression. When there are multiple input variables, we often refer to the method as multiple linear regression. Different techniques can be used to prepare or train the linear regression equation from data, the most common of which is called Ordinary Least Squares. It is common to therefore refer to a model prepared this way as Ordinary Least Squares Linear Regression or just Least Squares Regression [9].

### B. Gaussian Processes

Gaussian process regression is a nonparametric method based on modeling the observed responses of the different training data points (function values) as a multivariate normal random variable. For these function values, a priori distribution is assumed that guarantees smoothness properties of the function. Specifically, the correlation between two function values is high if the corresponding input vectors are close (in Euclidean distance sense) and decays as they go farther from each other. The posterior distribution of a to-be-predicted function value can then be obtained using the assumed prior distribution by applying simple probability manipulations [10].

### C. Multilayer Perceptron

A multilayer perceptron (MLP) is a deep, artificial neural network which is composed of more than one perceptron. They are composed of an input layer to receive the signal, an output layer that makes a decision or prediction about the input, and in between these two, are an arbitrary number of hidden layers that are the true computational engine of the MLP. MLPs with one hidden layer are capable of approximating any continuous function. A MLP model is often applied to supervised learning problems: they train on a set of input-output pairs and learn to model the correlation (or dependencies) between those inputs and outputs. Training involves adjusting the parameters, or the weights and biases, of the model in order to minimize error. Backpropagation is used to make those weight and bias adjustments relative to the error, and the error itself can be measured in a variety of ways, including by root mean squared error (RMSE) [11].

### D. Support Vector Regression

Support vector regression (SVR) is a successful method based on using a high-dimensional feature space (formed by transforming the original variables), and penalizing the ensuing complexity using a penalty term added to the error function. The model produced by SVR depends only on a subset of the training data, because the cost function for building the model ignores any training data close to the model prediction. The training vectors giving nonzero Lagrange multipliers are called support vectors. Non-support vectors do not contribute directly to the solution, and the number of support vectors is some measure of model complexity. This model is extended to the nonlinear case through the concept of kernel. A common kernel is the Gaussian kernel [12].

### E. K-Nearest Neighbor Regression

The K-nearest neighbor (KNN) regression method is a nonparametric method that bases its prediction on the target outputs of the K nearest neighbors of the given query point. Specifically, given a data point, the Euclidean distance between that point and all points in the training set is computed. Afterwards, the closest K training data points are picked and the prediction is set as the average of the target output values for these K points [13].

### F. Autoregressive Integrated Moving Average

Autoregressive integrated moving average (ARIMA) model is a generalization of an autoregressive moving average (ARMA) model. A stationary (time) series is one whose statistical properties such as the mean, variance and autocorrelation are all constant over time. ARIMA models are applied in cases where the data shows evidence of non-stationarity, therefore a differencing step can be applied a number of times to eliminate the non-stationarity [14].

### G. Vector Autoregression

Vector autoregression (VAR) is a stochastic process model used to capture the linear interdependencies among multiple time series. The VAR model extends the idea of univariate autoregression to **k** time series regressions, where the lagged values of all **k** series appear as regressors. VAR models generalize the univariate autoregressive model (AR model) such that in VAR we regress a vector of time series variables on lagged vectors of these variables [19].

# 3. METHODOLOGY

As described earlier, we first did a comparative study of different machine learning models available in Weka software [8] and then implemented two statistical models, ARIMA and VAR in Python. Once our models in python start producing accurate predictions, we plan to deploy these models in IoT based cloud computing platform. In this section, we will discuss the steps taken in detail.

### 3.1 Preprocessing Methods

Data preprocessing is an integral step in time series forecasting because the quality of the data and the usefulness of the information that can be derived from it directly affects the model performance when forecasting. For example, a large number of studies support the de-seasonalization of data possessing seasonality for neural networks [16, 17, 18]. Other preprocessing methods such as taking log difference and detrending have also been widely researched. The pre-processing methods are as follows:

1) No special preprocessing (Lag-Variables): The input variables to the machine learning model are lagged time series values, and the value to be predicted (target output) is the next value (for one-step ahead forecasting);
2) Time series differencing: We take the first backward difference and apply the forecasting model on this differenced series;
3) Seasonal Decomposition: As electricity consumption has a seasonal component, we used methods from python package `statsmodel` to decompose the time series.

Aside from these, we also performed data resampling and merging. The data was collected at a 15-minute interval; however, this data has a very high variance therefore we had to resample the data to multiple different intervals. Furthermore, because the Enertalk electricity consumption sensor [19] used in this study collects only the energy consumption data. Hence, we had to merge it with the weather data which we obtained from Open Weather Map API [20].

### 3.2 Experiment process

### A. Experiments on Kaggle Dataset using Weka

In the first step, we selected energy consumption datasets from Kaggle. Our goal was to find datasets that closely resembled the data we would eventually be forecasting. The datasets we selected are Hourly Energy Consumption (contained 12 datasets) [21], Germany electricity power for 2006-2017 [22], Electricity consumption history [23] and Solar generation and demand Italy 2015-2016 [24].

After selecting these datasets, we performed simple preprocessing and then converted the datasets to an Attribute-Relation File Format (.arff) for Weka to read. Once the dataset was converted into an .arff file, we loaded the dataset into Weka, selected the time series forecasting learner, specified the size of the test and training data and performed the forecasting using the 5 models we stated earlier. We varied the sizes of the test and training datasets and measured the resulting root mean square error, mean absolute error and relative absolute error (so we could compare the accuracy across different datasets).

The Figure 1 shows one of the graphs depicting the results of forecasting using SVM Regression method on the Electricity consumption history [22] dataset. (Note: the points in blue are the predicted values and the points in red are the actual values.)

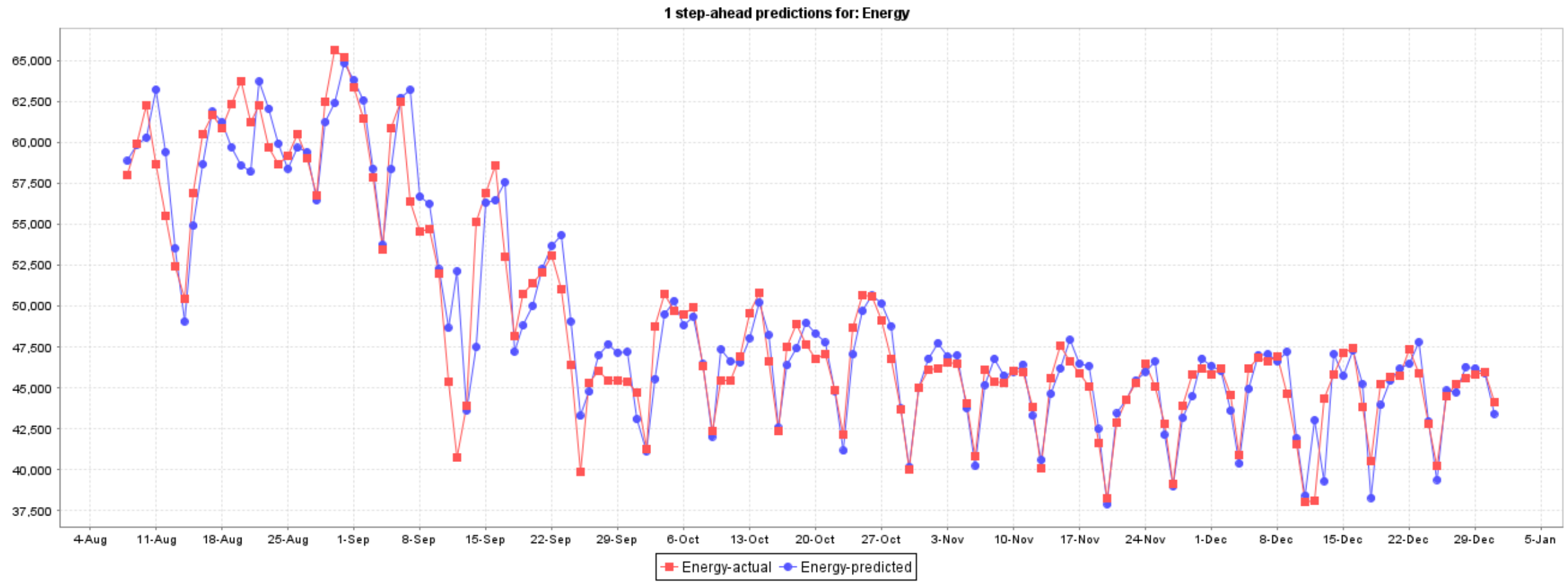

**Figure 1: SVM Regression Graph**

### B. Experiment on Korean household data

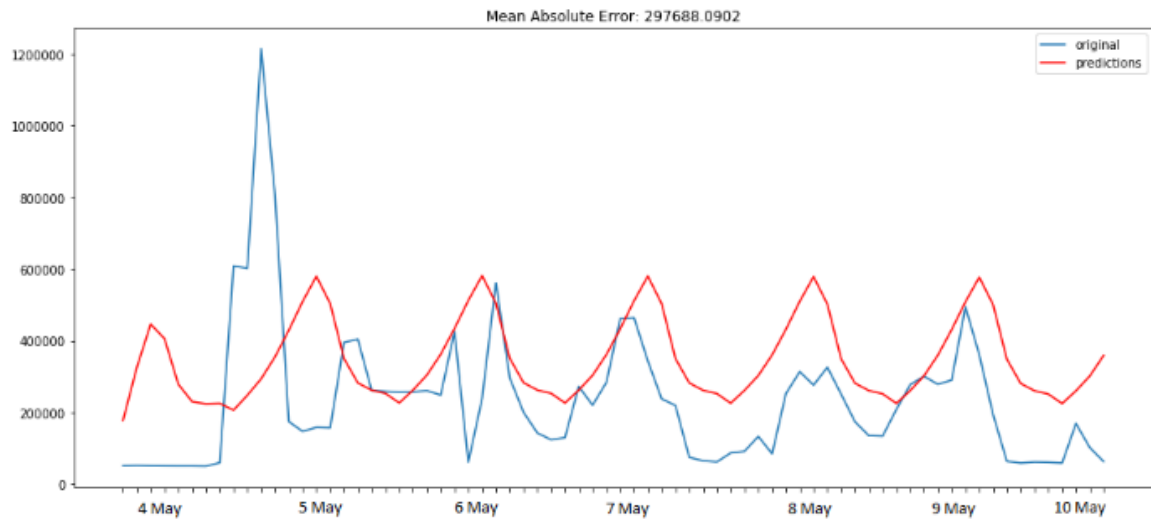

**Figure 2: ARIMA Hourly Graph – MAE:297688**

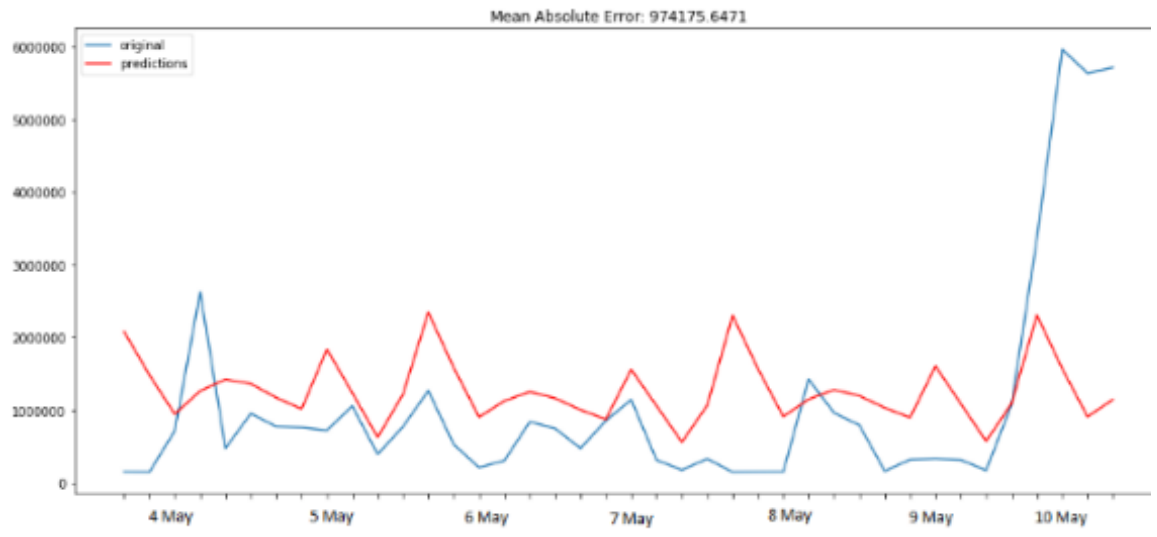

**Figure 3: ARIMA 3 Hourly Graph – MAE:974175**

The results from Weka which we will discuss in the next section show that the models rooted in statistics like SVM regression and Gaussian Processes regression performed better than other models therefore we decided to use statistical models for our python implementation. We determined that ARIMA would be a good model to predict univariate data and VAR would be a good option for multivariate data. We implemented both models using the python package `statsmodels`. The main problem we faced was that the size of the energy consumption data we collected was very short (from 15th December 2019-3rd May 2020). This resulted in a sub-optional performance of developed models. To improve the accuracy of the model we tried to experiment on data with a longer interval, so we resampled data from a 15 minutes interval to 1 hour, 3 hours, 6 hours, 12 hours and daily interval. The following are some graphs showing the forecasting energy consumption using ARIMA.

## 4. Result and Comment on Performance

The graphs shown in Fig. 6 – Fig. 15 represent the performance of each model on different datasets shown in Table 1. (Note: the black line represents Linear Regression, green represents Gaussian Processes, red represents Multilayer Perceptron, yellow represents SVM Regression, blue represents KNN Regression).

**Table 1: Kaggle datasets used in the experiments**

| | |
|---|---|
| Fig. 4 | Dataset containing energy consumption data for a single house from 2013–2016. |
| Fig. 5 | Data provided by Open Power System Data containing country wide electricity consumption of Germany from 2006–2017. |
| Fig. 6 | Data from ENTSO-E Transparency containing total electricity consumption in Italy during the years 2015-2016. |
| Fig. 7 | Dataset is a part of Hourly Energy Consumption datasets provided by the grid operator PJM Interconnection. It contains energy consumption recorded by Northern Illinois Hub (NI). |
| Fig. 8 | Dataset is also a part of the PJM Datasets. It contains consumption recorded by The Dayton Power and Light Company. |
| Fig. 9 | Dataset is a part of the PJM datasets. It consists of consumption recorded by Duke Energy Ohio. |
| Fig. 10 | Dataset is also a part of the PJM Datasets. It contains consumption recorded by Dominion Virginia Power. |
| Fig. 11 | This dataset was also a part of the PLM Datasets. It contains consumption recorded by Duquesne Light Co. |

We experimented on the Korean household data using Weka as well. The graph in Figure 14 shows the predictions using SVM Regression on the hourly and 3 hourly Korean household data. (Note: the points in blue are the predicted values and the points in red are the actual values. The unit of energy consumption is milliwatts.)

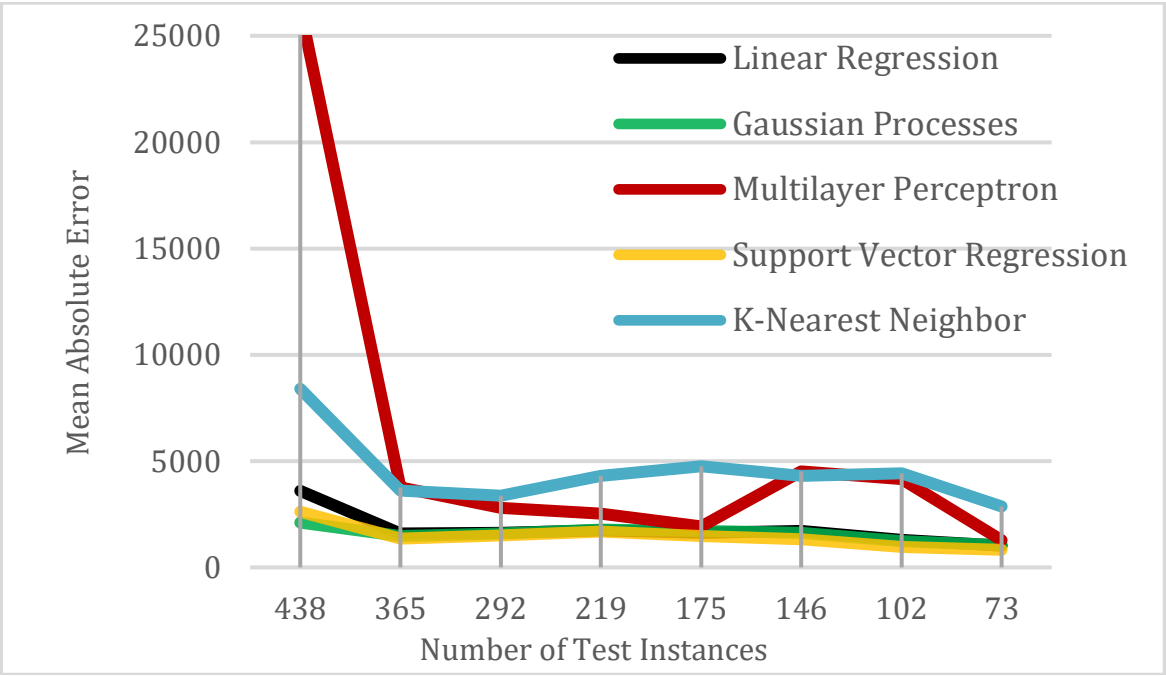

**Figure 4: Electricity Consumption History [22]**

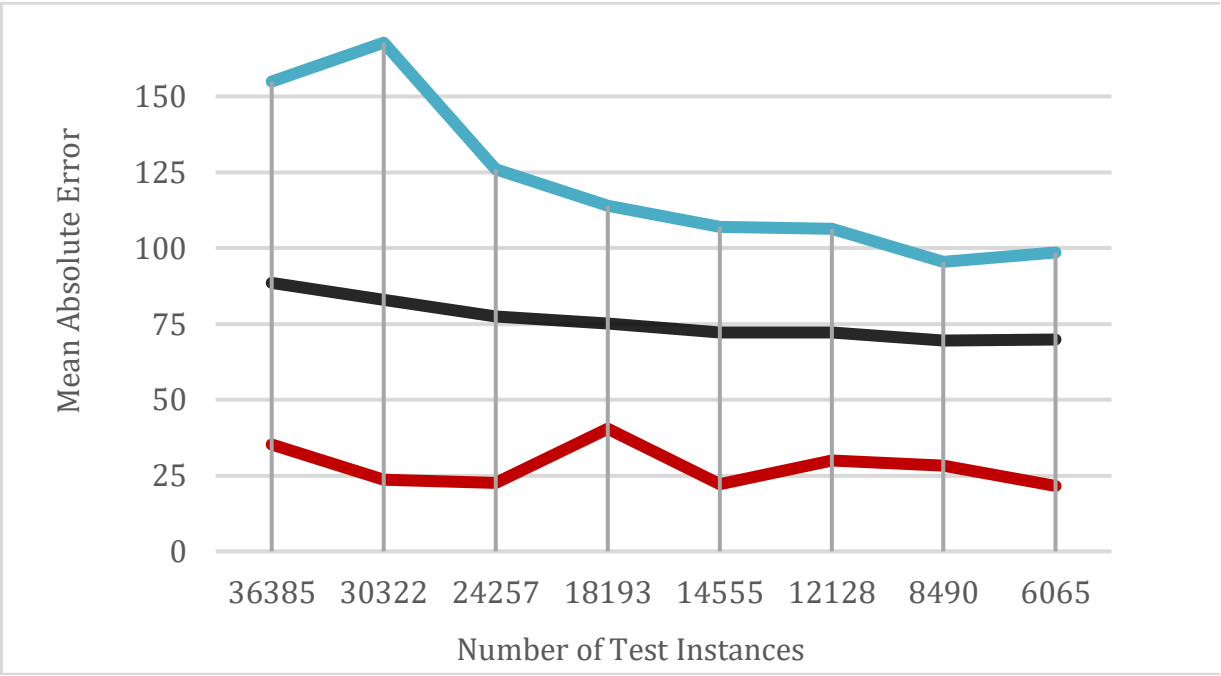

**Figure 8: Hourly Energy Consumption [20] - DAYTON**

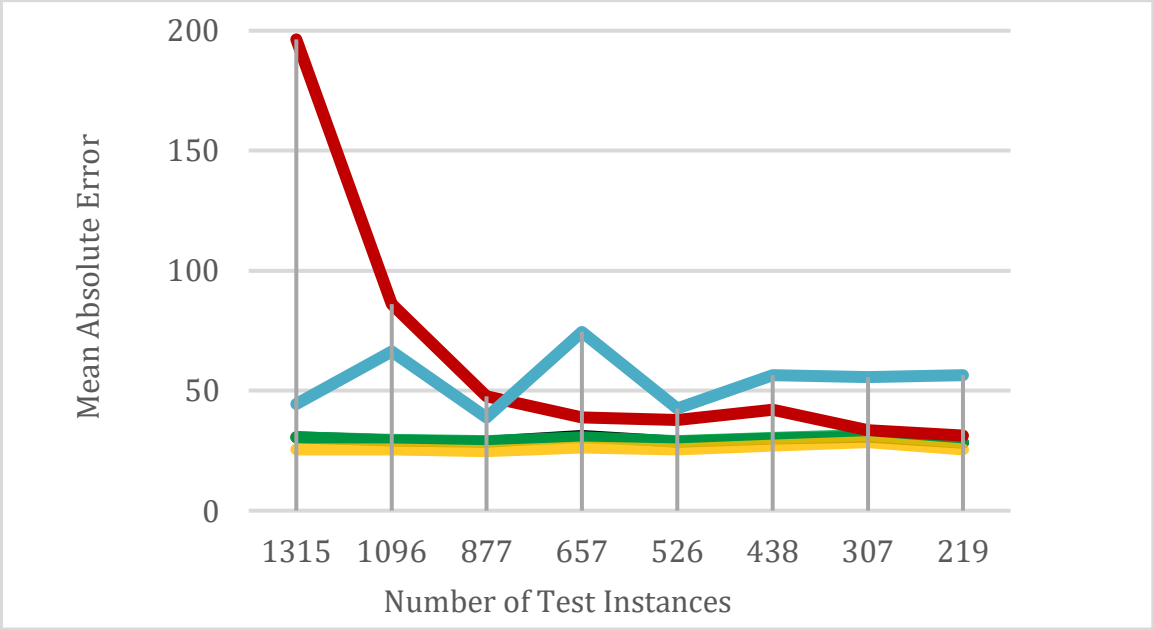

**Figure 5: Germany electricity power for 2006-2017 [21]**

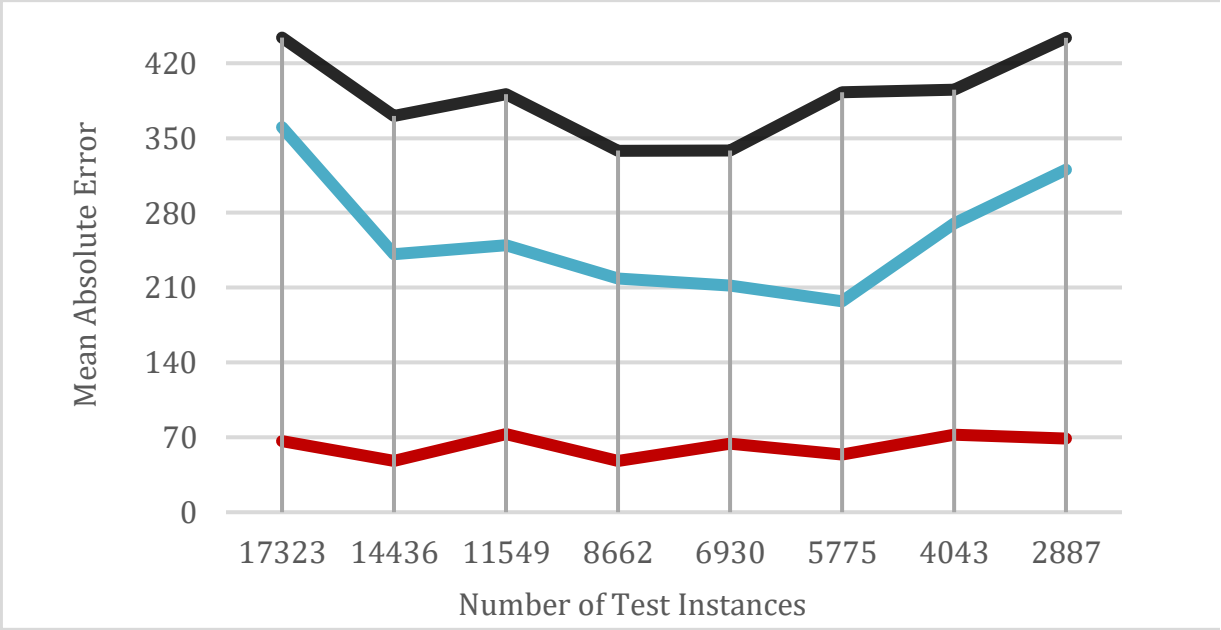

**Figure 9: Hourly Energy Consumption [20] - DEOK**

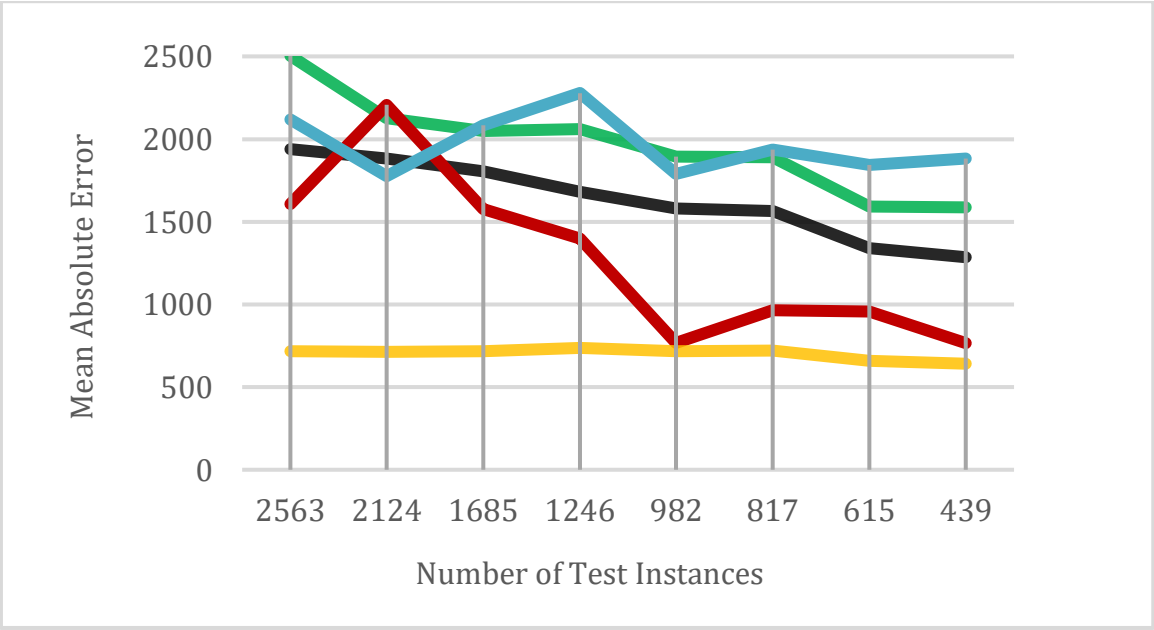

**Figure 6: Solar generation and demand Italy 2015-2016 [20]**

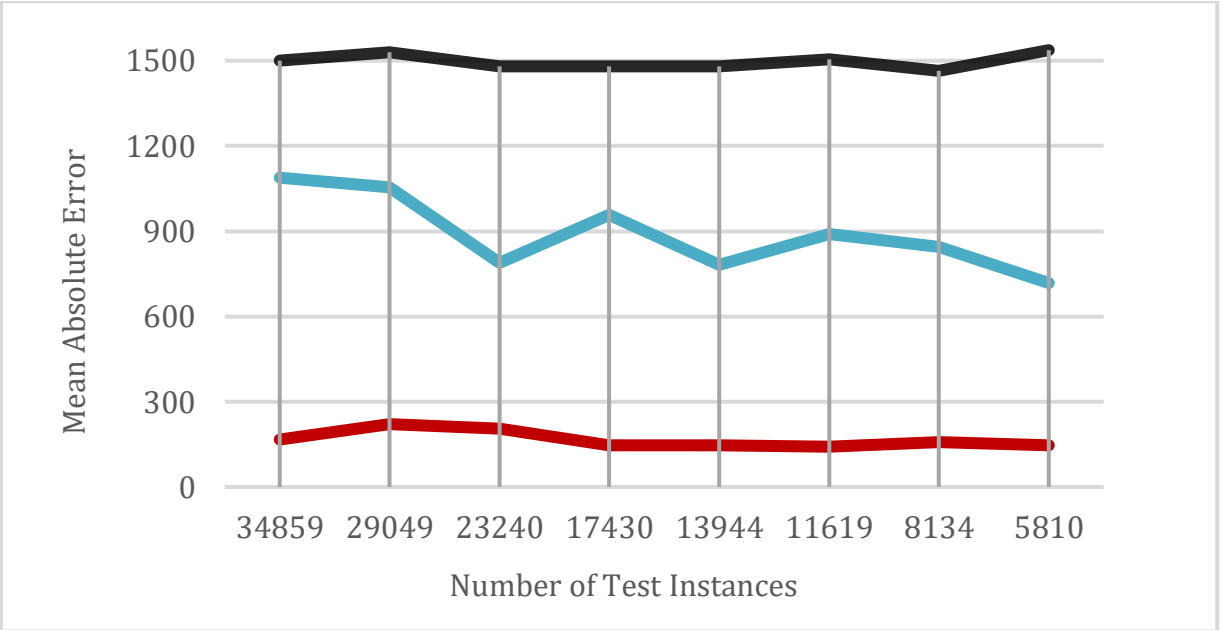

**Figure 10: Hourly Energy Consumption [20] - DOM**

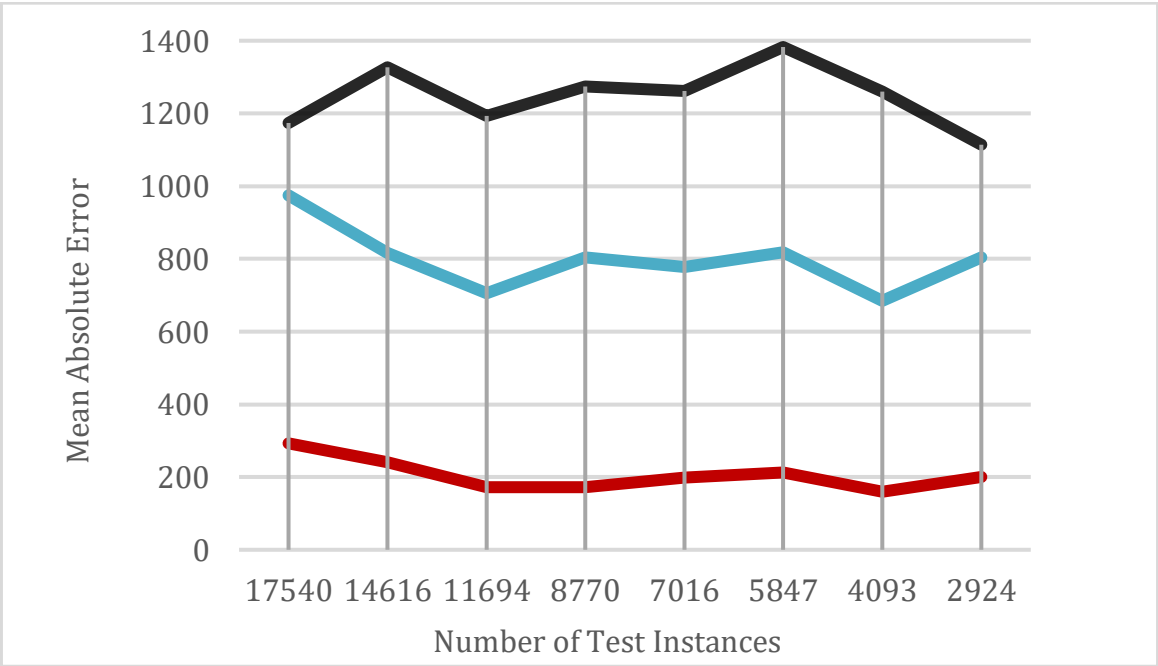

**Figure 7: Hourly Energy Consumption [20] - NI**

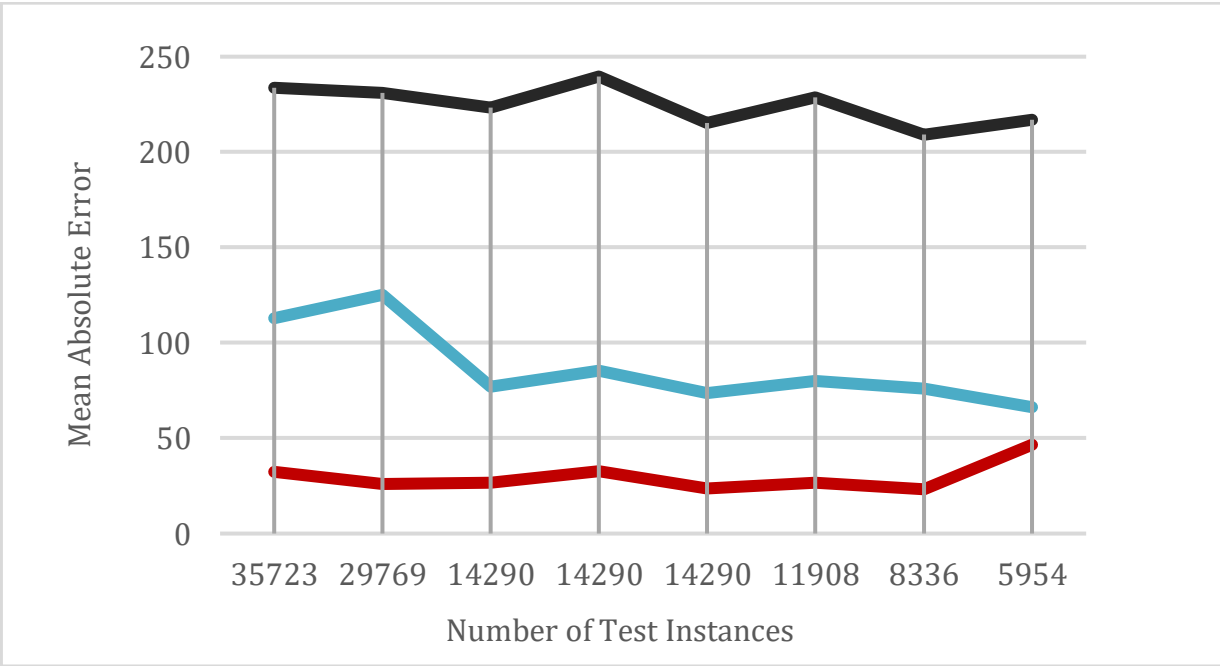

**Figure 11: Hourly Energy Consumption [20] - DUQ**

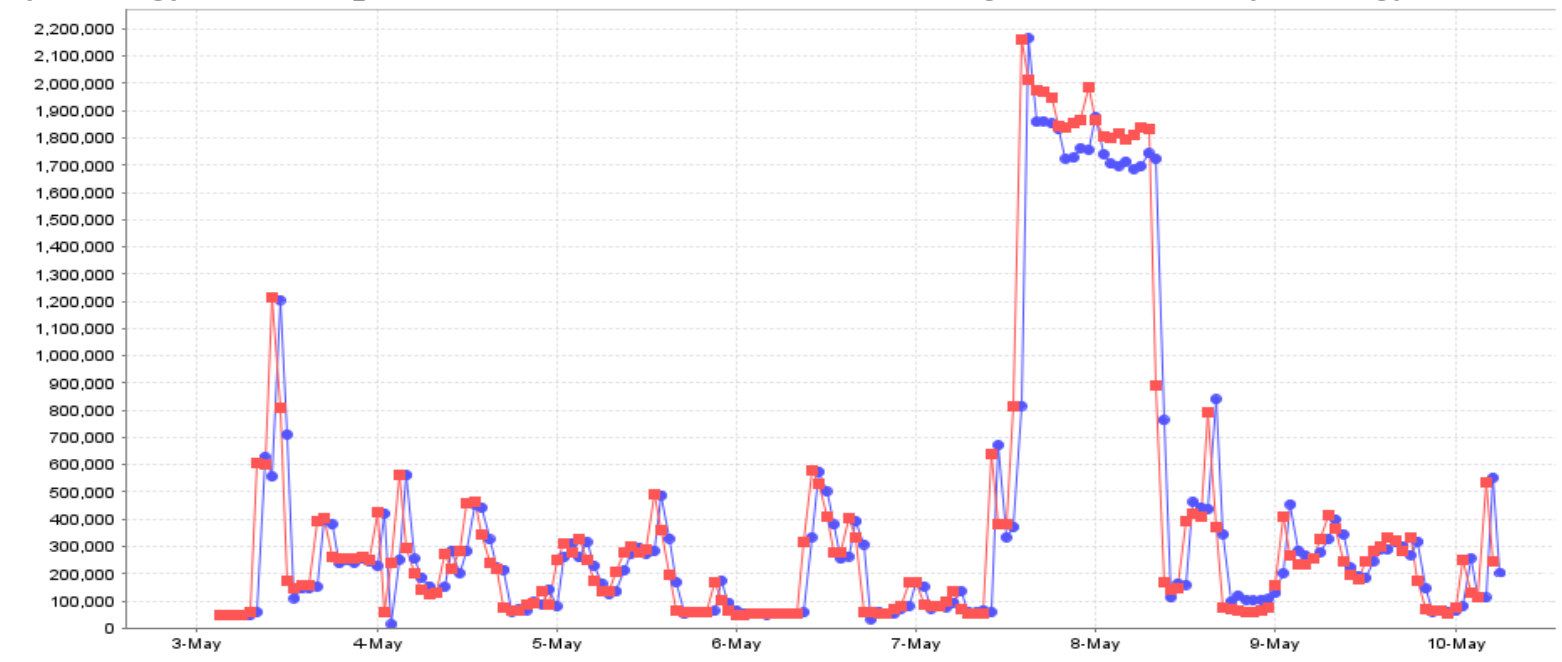

**Figure 12: SVM Regression Graph for hourly Korean household data**

From the presented results it is clear that overall support vector regression provides the best performance regardless of the size of the test and training dataset. The next best performance is provided by models Gaussian Processes and Multilayer Perceptron. For the forecasting with ARIMA, the datasets with 1 hour, 3 hours and 6 hours intervals were able to capture the trend. The VAR model required weather data to be merged with the energy consumption data and we didn't have the weather data for May 2020 thus the dataset for VAR was even smaller than ARIMA however it was still able to somewhat capture the trend for a day ahead prediction. Overall, the performance of both ARIMA and VAR models was worse than Weka models. We believe that this is caused by the fact that the dataset we used to train the ARIMA and VAR models was much smaller.

# 5. Conclusion

The performance of five classical machine learning models for time series forecasting on household energy consumption was investigated. We analyzed the model's performance while varying the sizes of the test and training datasets and across 15 different datasets. Our study showed that SVM Regression offered the best performance while Multilayer Perceptron, and Gaussian Processes also offered next best performances for time series forecasting on household energy consumption. After the comparison study we implemented ARIMA and VAR models in python to forecast the energy consumption in South Korean households. While these models were able to capture the trends, they were not as accurate as what we had expected. Our next objectives are to: 1) Integrate the models we implemented in python in an energy prediction platform implemented using Google Cloud Platform (GCP) so that we can get energy predictions in real time; and 2) Investigate use of RNN model for time series forecasting and compare performance with ARIMA and VAR models.

# Acknowledgement

This work is funded by Incheon Techno Park, Songdo-dong, Yeonsu-gu, Incheon under the project *Development of IoT devices for Elderly to Provide Improved Access to Digital Information* (독거노인의 디지털정보 접근성 향상을 고려한 디바이스 개발) in collaboration with Ninewatt Inc. (http://www.ninewatt.com/).